\documentclass[letterpaper, 10 pt, conference]{ieeeconf}  

\IEEEoverridecommandlockouts                              

\usepackage{amsmath}
\usepackage{amssymb}
\usepackage{color}
\usepackage{graphicx}

\usepackage{url}

\usepackage{soul}

\renewcommand{\baselinestretch}{0.95}

\title{\LARGE \bf
Map-Predictive Motion Planning in Unknown Environments
}

\author{Amine Elhafsi$^{1}$, Boris Ivanovic$^{1}$, Lucas Janson$^{2}$, Marco Pavone$^{1}$
\thanks{*This work was supported in part by NSF, Award Number: 1931815, by NASA under the NSTRF program, and by TRI. This article solely reflects the opinions and conclusions of its authors and not of NSF, NASA, TRI or any other Toyota entity.}
\thanks{$^{1}$Department of Aeronautics and Astronautics, Stanford University, Stanford, CA 94305, USA
        {\tt\small \{amine, borisi, pavone\}@stanford.edu}}%
\thanks{$^{2}$Department of Statistics, Harvard University, Cambridge, MA 02138, USA
{\tt\small ljanson@fas.harvard.edu}}%
}

\begin{document}

\maketitle
\thispagestyle{empty}
\pagestyle{empty}

\begin{abstract}
Algorithms for motion planning in unknown environments are generally limited in their ability to reason about the structure of the unobserved environment. As such, current methods generally navigate unknown environments by relying on heuristic methods to choose intermediate objectives along frontiers. We present a unified method that combines map prediction and motion planning for safe, time-efficient autonomous navigation of unknown environments by dynamically-constrained robots. We propose a data-driven method for predicting the map of the unobserved environment, using the robot's observations of its surroundings as context. These map predictions are then used to plan trajectories from the robot's position to the goal without requiring frontier selection. We demonstrate that our map-predictive motion planning strategy yields a substantial improvement in trajectory time over a na\"{i}ve frontier pursuit method and demonstrates similar performance to methods using more sophisticated frontier selection heuristics with significantly shorter computation time.
\end{abstract}

\section{Introduction}

As robots move from meticulously-organized research labs and factory floors into natural or human-built environments, they must be capable of safely and efficiently planning trajectories through unexplored spaces. This is particularly important for a variety of robotic applications, such as search and rescue operations in uncharted areas, autonomous driving with occluded perception, and planetary exploration.

\begin{figure}[t]
  \centering
  \includegraphics[width=0.85\linewidth]{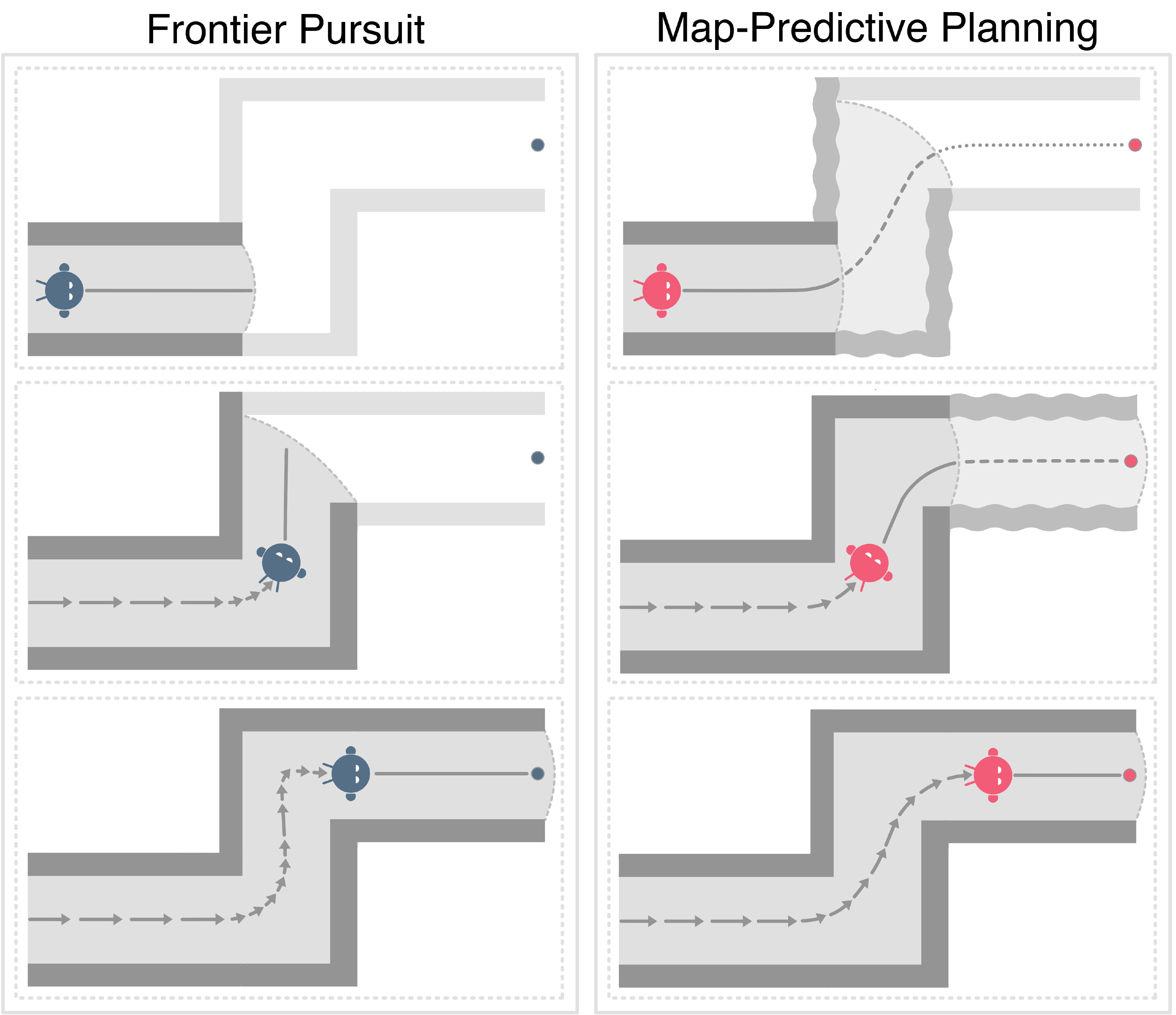}
  \caption{\textbf{Left:} A frontier pursuit method plans without reasoning about the unobserved environment. As a result, it is not prepared for the turn and must slow down excessively and turn sharply. \textbf{Right:} Our motion planning method accounts for potential upcoming turns via map prediction, enabling a smooth and more efficient trajectory through the hallway.}
  \label{hero}
  \vspace{-0.5cm}
\end{figure}

A common approach to planning in unknown environments relies on repeatedly replanning trajectories towards intermediate objectives within the robot's known environment, with the expectation that the robot will eventually arrive at its desired goal. However, such a planning strategy is shortsighted in that it does not reason about the map beyond the observed environment, resulting in greedy and inefficient trajectories. For example, a robot navigating an indoor environment may suddenly observe a perpendicular corridor, calling for an abrupt maneuver to complete the turn---especially if traveling at higher speeds as conceptualized by Fig.~\ref{hero}. Similarly, the pursuit of a frontier point, without reasoning about what may lie beyond, may lead the robot towards an occluded obstacle, requiring the execution of some emergency maneuver.

We begin by enumerating four desiderata of a motion-planning algorithm for unknown environments. First, the method must be capable of performing planning for robots with dynamic constraints. Second, robot safety must be guaranteed with respect to observed obstacles and the unobserved environment, where unseen obstacles may reside. The definition of safety in this work requires that the algorithm does not produce trajectories in collision with these regions of the workspace. Additionally, the robot should not put the robot into a state where future collisions may be unavoidable. Third, this method should be capable of incorporating reasoning about the unobserved environment in a manner that is invariant to the prediction model. Finally, the algorithm must lend itself to rapid replanning such that it is amenable to real-world applications. We motivate our proposed method by considering these desiderata among prior work related to this problem.

\section{Related Work}
Traditionally, motion planning strategies in unknown environments have relied on repeatedly replanning trajectories towards intermediate objectives until the final goal region is reached~\cite{BekrisKavraki2007,PivtoraikoMellingerEtAl2013,LiuWattersonEtAl2016}. This receding-horizon style of planning usually relies on unsophisticated heuristics to choose an intermediate objective (e.g. pursuing the frontier point nearest the goal at each iteration \cite{Yamauchi1997}; more sophisticated heuristics may be found in \cite{BurgardMoorsEtAl2005}). However, these heuristics often yield greedy, short-sighted behaviors as previously described.

Several works have proposed techniques for reasoning about the unobserved environment to navigate through them more efficiently. One class of works seek to use the robot's current knowledge of the environment to better inform high-level routing decisions through topologically complex environments (e.g. indoor building floor plans). \cite{SteinBradleyEtAl2018} proposes a method of navigating partially-revealed environments in minimum distance by modeling the problem as a Partially Observable Markov Decision Process (POMDP), where optimal actions correspond to frontiers that lead directly to the goal rather than lengthy detours or dead ends. Towards a similar objective, \cite{CaleyLawranceEtAl2019} passes an occupancy grid encoding a robot's current knowledge of the environment through a Convolutional Neural Network (CNN) and uses the output to weigh frontiers based on their likelihood of leading to a point of interest. While these works enable shorter-distance navigation through unknown environments to a desired goal, they crucially do not consider robot dynamics. 

Another class of such works look to directly predict the structure of the unknown environment. \cite{AydemirJensfeltEtAl2012} represents building floor plans as graphs, where nodes correspond to rooms and edges indicate traversable paths, and proposes a method of predicting extensions to the known environment topology. Such an approach would be useful for improving high-level routing, yet the graph representation of the environment is not amenable to low-level trajectory planning for dynamic robots. \cite{ChangLeeEtAl2007b, StroemNenciEtAl2015} more explicitly attempt to predict an occupancy grid of the unknown environment by comparing the robot's current observations against a collection of previously stored maps. The prediction process compares the surroundings of an unexplored region with the built map of explored regions. However, the accuracy and feasibility of this prediction depends on the robot having previously observed similar structure, hampering generalization. Drawing inspiration from image inpainting literature (e.g. \cite{LiuRedaEtAl2018, YuLinEtAl2018}), \cite{ShresthaTianEtAl2019, KatyalPopekEtAl2019} treat the problem of occupancy grid prediction as one of image completion. These works propose using CNNs which take as input an occupancy grid of the observed environment and produce occupancy predictions for the unknown regions as outputs. However, a CNN-based approach requires that the inputs and outputs be of some fixed sizes set at training time. Due to the dynamic nature of navigation in unknown environments, available information and desired outputs may vary. For example, in confined environments a robot's sensing is limited and only a small subset of the CNN input would be required whereas sparse environments would require much larger inputs to provide enough context. Similarly, high-speed navigation would require a significantly larger prediction region than low-speed travel. 

Several works have also sought to combine reasoning about the unknown environment with planning. \cite{GuptaDavidsonEtAl2017} proposes a neural network architecture that is trained to learn a mapping from first-person views to a discrete movement action set (i.e. stay still, move forward, turn left/right), and is capable of navigating novel environments. Yet, beyond the limited action space, such an end-to-end learning-based approach is effectively a black-box function which provides little insight behind the robot's reasoning for choosing a particular action. \cite{RichterVega2018} introduces a planner for dynamic, high-speed navigation through unknown environments. The planning problem is formulated as a POMDP where actions are chosen from a library of motion primitives minimizing time and collision probability. Reasoning about the unobserved environment is implicitly performed via the prediction of collision probabilities which are learned as a function of the robot action and a set of hand-coded features encoding the robot's immediate observations. \cite{RichterRoy2017} proposes a similar planner for visual navigation by mapping images and robot actions to collision probabilities using a neural network. While these works demonstrate significant performance gains in terms of trajectory time, they crucially lack formal safety guarantees.

Along the lines of combining map prediction and planning, \cite{JansonHuEtAl2018} introduces a general framework for tackling this problem and is closely related to this work. As such, we formulate our motion problem according to this framework and experimentally compare to a baseline algorithm which is an instantiation of this framework.

\emph{Statement of Contributions:} To address the problem of motion planning in unknown environments, we make the following three contributions:
(1) We introduce a data-driven approach to predicting an environment's structure in regions that are occluded or beyond sensor range.
In particular, our method produces human-interpretable probabilistic map predictions that can be passed to any general motion planning algorithm.
(2) We propose a method of incorporating these predictions within a motion planning framework that guarantees robot safety. This framework makes use of the predictions to penalize trajectories passing through regions with a high probability of occupancy.
(3) We demonstrate, with extensive numerical experiments, that using these predictions yields a substantial improvement in trajectory time over a na\"{i}ve frontier pursuit method and significant computation time reduction over methods using more sophisticated frontier selection heuristics. Our method uniquely addresses all four aforementioned desiderata; it is capable of planning safe, dynamically-constrained trajectories with low computation times amenable to real-time performance.


\section{Problem Formulation}
In order to more efficiently navigate unknown environments, we seek to predict the probability of occupancy at some set of unobserved points in the environment given some nearby observations as context. We then formulate a motion planning problem seeking a safe trajectory minimizing a desired cost function through these environments.

\subsection{Map Prediction}
We begin with the assumption that robot perception is achieved via some range-limited line-of-sight mechanism such as a lidar. We further assume that the robot's perception is incorporated within a deterministic occupancy grid representation of the environment, encoding regions of free, occupied and unknown space. Finally, we make the assumption that the environment is static. 

We consider the environment's occupancy grid to be an $n$-tuple random variable $(Y_1,\ldots,Y_n)$ whose elements represent the occupancy at grid cell $i$ and are described by some unknown distribution. We denote $y_i \in \mathcal{Y} = \{0,1\}$ to be a realization of $Y_i$. We also define a map's set of spatial coordinates
as the set $\mathcal{X} \in \mathbb{R}^2$, with the observed and unobserved regions denoted by $\mathcal{X}_{obs} \subset \mathcal{X}$ and $\mathcal{X}_{un} = \mathcal{X} \setminus \mathcal{X}_{obs}$, respectively. We let the context set $\mathcal{C}=\{(x_i, y_i)\}_{i=1:c}$,
correspond to the set of $c$ observed locations $x_j \in \mathcal{X}_{obs}$ for $j = 1, \ldots, c$ paired with their known occupancy state $y_i$. Finally, the target set is denoted by $\mathcal{T} = \{x_i\}_{i = (c+1):(c+t)}$ and corresponds to the $t$ spatial coordinates $x_k \in \mathcal{X}_{un}$ for $k = c+1, \ldots, c+t$ for which we seek the occupancy value.

It is then assumed that there exists some stochastic process $P$ over functions $f: \mathcal{X} \rightarrow \mathcal{Y}$. For some function $f \sim P$, we set $y_i = f(x_i)$. As such, the stochastic process defines the joint distribution $P(\{f(x_i)\}_{i=1:(c+t)})$ and thus the conditional distribution $P(f(\mathcal{T}) \mid \mathcal{C}, \mathcal{T})$. Given this formulation, the objective is to determine this conditional distribution.

\subsection{Motion Planning with Map Predictions} \label{SectionIIIb} 
The motion planning problem is concerned with planning trajectories to minimize some objective function of a trajectory $\sigma$, subject to initial and terminal boundary conditions, while satisfying dynamic, control and safety constraints. We define a trajectory, $\sigma$, as a tuple of states, \( \textbf{s} \left( t \right)  \), control inputs, \( \textbf{u} \left( t \right) ,  \) and duration, \( T \). In order to make use of map predictions, we focus on strategies that repeatedly replan trajectories as new information is observed. However, rather than computing partial trajectories to intermediate objectives at each iteration, we choose to plan all the way to the goal using predictions to guide the trajectory computation in unknown regions following the framework presented by \cite{JansonHuEtAl2018}. In effect, at each iteration we seek trajectories solving the optimization problem:
\begin{align}
    \underset{\sigma_\text{k}, \sigma_\text{u}}{\text{minimize}} \hspace{1cm} & c(\sigma_\text{k}, \sigma_\text{u}) + h(\sigma_\text{u}, \Phi) \nonumber\\
    \text{subject to} \hspace{1cm} & \mathbf{s}_\text{k}(t_\text{initial}) = s_\text{initial} \nonumber\\
    & \mathbf{s}_\text{k}(t_\text{frontier}) = \mathbf{s}_\text{u}(t_\text{frontier}) \nonumber\\
    & \mathbf{s}_\text{u}(t_\text{final}) \in \mathcal{S}_\text{final} \nonumber\\ 
    & \sigma_\text{k}, \sigma_\text{u} \in \Sigma_\mathcal{D} \label{MPformulation}\\
    & \mathbf{u}_\text{k}(t), \mathbf{u}_\text{u}(t) \in \mathcal{U} \ \forall\ t \nonumber\\
    & \mathbf{u}_\text{k}(t_\text{frontier}) = \mathbf{u}_\text{u}(t_\text{frontier}) \nonumber\\
    & \mathbf{s}_\text{k}(t) \in \mathcal{S}_\text{free} \ \forall\ t \nonumber\\
    & \mathbf{s}_\text{k}(t) \notin \mathcal{S}_\text{unsafe} \ \forall\ t. \nonumber
\end{align}
In this formulation, we partition the full trajectory $\sigma$ into subtrajectories $\sigma_\text{k}$ and $\sigma_\text{u}$. We designate $\sigma_\text{k}$ as the immediate trajectory in the known region for the robot to execute, terminating at a frontier. We use $\sigma_\text{u}$ to represent a tentative future trajectory in the unknown region beginning from the end point of $\sigma_\text{k}$. The objective function consists of the terms $c(\sigma_\text{k}, \sigma_\text{u})$ the primary cost we wish to minimize, and $h(\sigma_\text{u}, \phi)$, a penalty term to discourage $\sigma_\text{u}$ from passing through regions of high predicted occupancy. States and controls share the same subscript convention. We enforce initial and final state constraints with $s_\text{initial}$ representing the initial robot state and $\mathcal{S}_\text{final}$ representing the set of acceptable goal states. The set of dynamically feasible trajectories and admissible control actions are denoted $\Sigma_\mathcal{D}$ and $\mathcal{U}$, respectively. Finally, $\mathcal{S}_\text{free}$ represents the set of collision-free states while $\mathcal{S}_\text{unsafe}$ represents any chosen set of states that compromise the robot's safety, such as Inevitable Collision States (ICS)~\cite{FraichardAsama2004}. Within this setting, an ICS is defined as a state for which the robot will either eventually collide with an obstacle or enter unobserved space (where obstacles may reside), despite any control sequence the robot may execute. Note that state and control continuity are enforced at $t_\text{frontier}$, the time at which the trajectory first crosses a frontier into unknown space. The last two constraints guarantee safety by requiring $\sigma_\text{k}$ to be collision free and outside of any ICS. 

Rather than thresholding predictions and enforcing that $\sigma_\text{u}$ be collision free with respect to the predicted map, our formulation seeks to simply bias $\sigma_\text{u}$ away from regions of high occupancy probability. We find that this formulation is more forgiving in the event that predictions are inaccurate, e.g. in cases where there is only one route to a goal and inaccurate thresholded predictions would wall it off, leading to infeasibility. In practice, we solve this problem at each iteration, execute a portion of $\sigma_\text{k}$, update the robot's map of the environment, and repeat this process until the goal is reached. Although the robot should never execute any part of $\sigma_\text{u}$, its presence in the optimization influences $\sigma_\text{k}$ to navigate the known environment in a more informed manner, e.g., better situating the robot for a predicted turn in the future.

\section{Map-Predictive Motion Planning with\\Conditional Neural Processes}

\subsection{Map Prediction}

  \begin{figure}[t]
      \centering
      \includegraphics[width=0.8\linewidth]{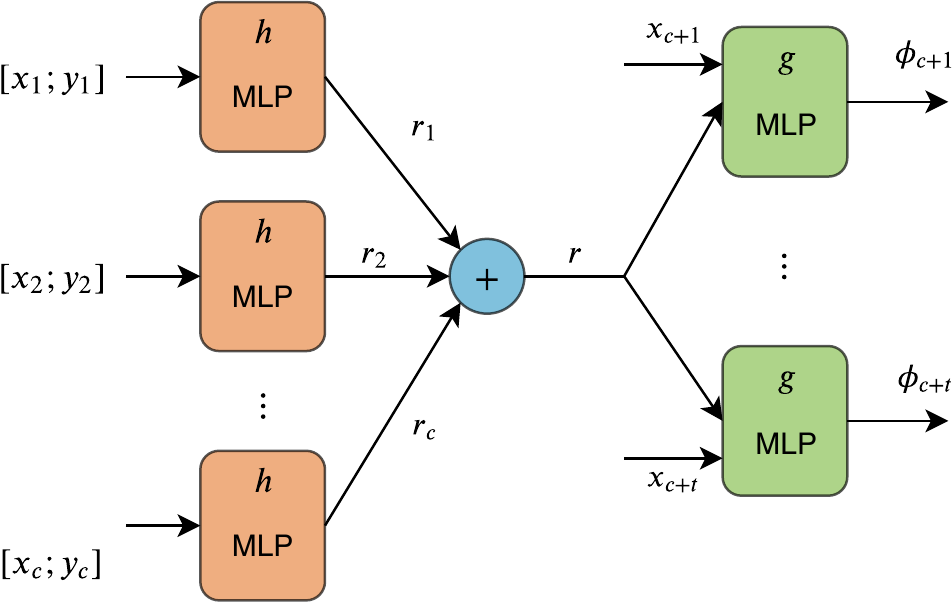}
      \caption{Our map prediction CNP architecture. Context points and their occupancies $\{x_i, y_i\}_{i=1:c}$ are first encoded by $h$, a multilayer perceptron (MLP), to generate representation vectors $\{r_i\}_{i=1:c}$. These representation vectors are then averaged to produce an overall scene encoding $r$. The vector $r$ is then concatenated with each of the target points $\{x_i\}_{i = (c+1):(c+t)}$ and fed through our decoder $g$, another MLP, which produces occupancy probabilities $\{\phi_i\}_{i = (c+1):(c+t)}$ for each of the target points.}
      \label{architecture}
      \vspace{-0.5cm}
  \end{figure}
  
Machine learning-based approaches are uniquely primed to model the distribution over maps, especially given the variety and complexity of real-world environments. Machine learning models can implicitly capture the salient features of environments from data, rather than formulating their distribution explicitly. Specifically, we choose a Conditional Neural Process (CNP)~\cite{GarneloRosenbaumEtAl2018} architecture to directly parametrize the conditional stochastic process approximating~$P(f(\mathcal{T}) \mid \mathcal{C}, \mathcal{T})$. In contrast to CNNs, CNPs are flexible in that they accept arbitrarily many inputs, as determined by the size of $\mathcal{C}$, as well as query arbitrarily many predictions, as determined by $\mathcal{T}$. This approach also has similarities to Gaussian process regression, yet is more scalable as it overcomes the need for costly matrix inversions.

The CNP architecture consists of an encoding neural network, an aggregation operation and a decoding neural network to produce parameters of the approximating conditional distribution $Q(f(\mathcal{T}) \mid \mathcal{C},\mathcal{T})$. The CNP architecture is illustrated in Fig.~\ref{architecture}.

The encoding procedure involves using the encoding neural network, $h(x_i, y_i)$, to produce embeddings $r_i$ for each of the $c$ context pairs $(x_i, y_i) \in \mathcal{C}$. These embeddings are then aggregated into a single conditioning representation vector $r$. In the CNP formulation, this aggregation operation, $a(r_{1:c})$, may take the form of any commutative operation mapping multiple vectors in $\mathbb{R}^d$ to a single vector in $\mathbb{R}^d$. Finally, the decoding network, $g(x_i, r)$ produces a vector of parameters, $\phi$, of a distribution over the occupancy of some $x_i \in \mathcal{T}$.

The encoding network was designed as a four-layer fully-connected feedforward network to produce a 256-dimensional $r_i$. Each layer consists of 256 neurons with Rectified Linear Unit (ReLU) activations. For the aggregation operation, we choose to simply average the embeddings since this operation weighs all information equally and ensures a similar magnitude between the embeddings and $r$---characteristics beneficial to network stability once deployed.

Finally, the decoding network consists of another four-layer fully-connected network with each layer consisting of 256 ReLU activated neurons. The final layer feeds into a sigmoid output representing a scalar-valued $\phi$. The model was trained to minimize a negative log-likelihood loss function such that $\phi$ could be interpreted as parametrizing a Bernoulli distribution over a target point's occupancy. With the predicted occupancy values in hand, we now describe how we these predictions are incorporated within our motion planning framework.

\subsection{Motion Planning} \label{MotionPlanningIIIB}
For this work, we are interested in navigating unknown environments in minimum-time and as such take $c(\sigma_\text{k}, \sigma_\text{u})$ to represent the overall trajectory duration, $T$. We also choose
\begin{equation*}
    h(\sigma_\text{u}, \Phi) = \alpha \int_{\mathbf{s}_\text{u}} \frac{1}{1 - \Phi(\mathbf{s}_\text{u}) + \epsilon}\ d\mathbf{s}_\text{u}
\end{equation*}
which yields a higher cost for $\sigma_\text{u}$ passing through regions of high occupancy probability. Here, $\Phi(\mathbf{s}_\text{u})$ is a function mapping the state $\mathbf{s}_\text{u}$ to a predicted occupancy,  $\alpha$ is a scaling parameter chosen to be 0.25 experimentally and $\epsilon$ is a small, positive constant introduced to avoid singularities. In this work we present one possible selection of $c$ and $h$, but we emphasize that the framework is general and alternative choices may be selected depending on designer preferences.

We take a simple friction circle car as our system of interest with dynamics 
\begin{equation*}
    m\mathbf{\ddot{s}(t)} = 
    \begin{bmatrix} 
        \cos{\theta(t)} & -\sin{\theta(t)} \\
        \sin{\theta(t)} & \cos{\theta(t)}
    \end{bmatrix}
     \mathbf{u}(t)
\end{equation*}
where $m$ is the car's mass, $\theta$ is the car's instantaneous heading. The state $\mathbf{s} \in \mathbb{R}^2$ represents the robot position. The control input $\mathbf{u}$ may be explicitly written as $[\mathbf{u}_\text{long}\ \mathbf{u}_\text{lat}]^\intercal$ with the first and second components respectively representing longitudinal and lateral forces in the vehicle body frame. The friction circle constraint is expressed as 
\begin{equation*}
    \|\mathbf{u}\| \leq \mu m g
\end{equation*}
where $\mu$ is the friction coefficient and $g$ is the gravitational constant. A constraint specifying the minimum turning radius, $R_\text{min}$, is also enforced as 
\begin{equation*}
    \mathbf{u}_\text{lat} \leq m \frac{\|\mathbf{\dot{s}}\|^2}{R_\text{min}}.
\end{equation*}
We modeled our system after a radio-controlled (RC) car with $m =$ 2.5 kg, $\mu =$ 0.9, $R_{\text{min}}$ = 0.5 m.

  \begin{figure*}[ht]
    \centering
    \includegraphics[width=0.30\linewidth]{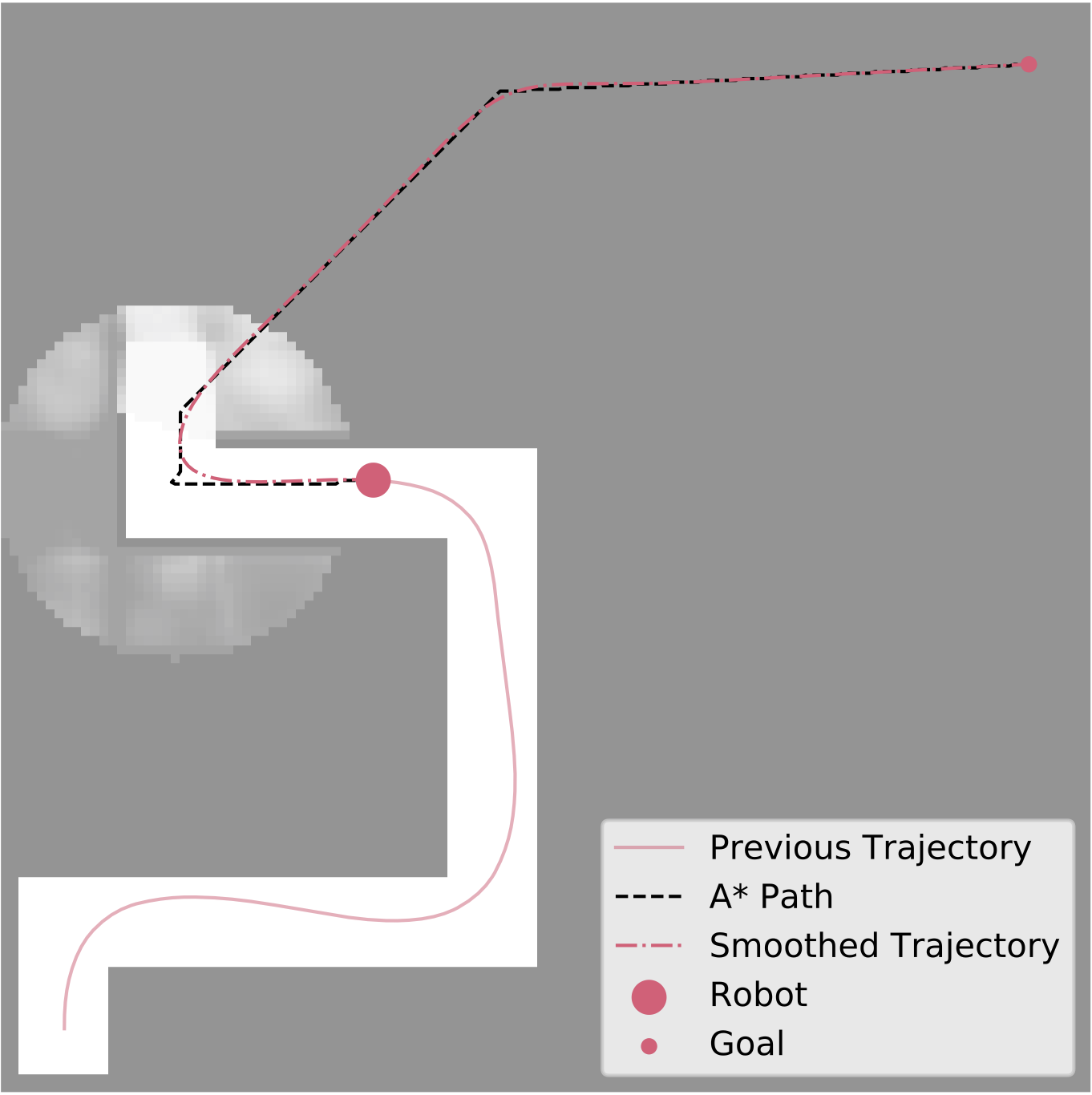}
    \includegraphics[width=0.30\linewidth]{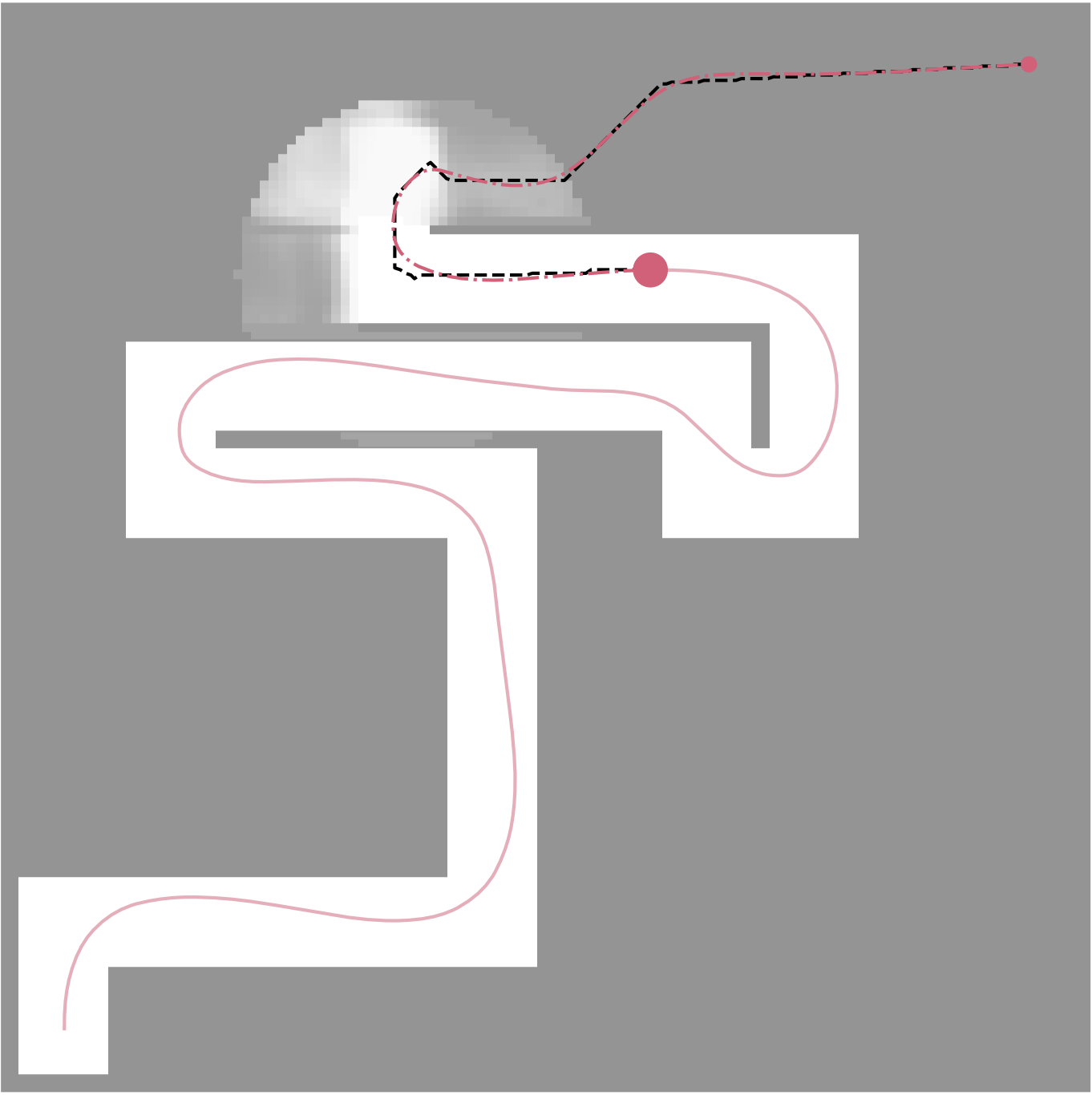}
    \includegraphics[width=0.30\linewidth]{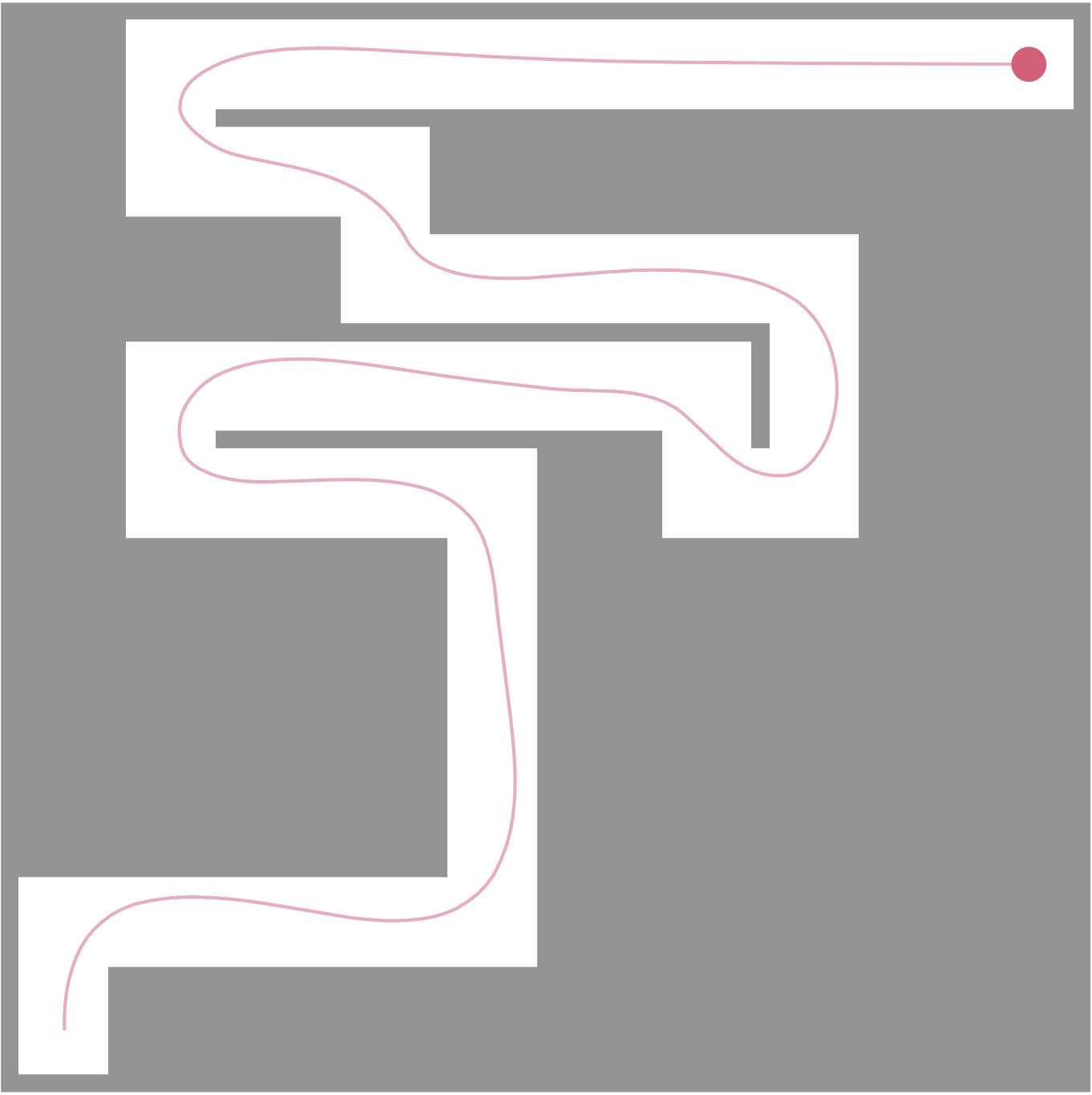}
    \caption{\textbf{Left:} Our method plans from the robot's current position to the goal, taking into account the predicted occupancy probabilities outside of the known environment. Our CNP's predictions are visualized centered at the current frontier point. Occupancy probabilities range from white (free space) to dark grey (occupied). \textbf{Middle:} A* plans through a higher occupancy probability region, even though our model correctly predicted a hallway to the left. This occurs because our A* heuristic trades off a higher occupancy probability penalty for a lower distance to the goal.
    \textbf{Right:} This is mitigated by our algorithm's frequent replanning; there is no detriment to the executed trajectory.
    }
    \label{pred_vis}
    \vspace{-0.5cm}
  \end{figure*}

To approximate solutions to \eqref{MPformulation} at each iteration, we perform a grid search to find a reference path from the robot's position to the goal, smooth this path to be feasible under our system's dynamics and then optimize the speed profile for minimum-time traversal. As compared to \cite{JansonHuEtAl2018}'s use of the sampling-based planner FMT* \cite{JansonSchmerlingEtAl2015}, we choose this decomposed motion planning strategy with practical realizability (i.e. real-time computation requirements) in mind.

We use the A* search algorithm to find the path from the robot's position to the goal. We use the Euclidean distance heuristic and treat obstacles as impenetrable within the known region. Within the unknown region, we attempt to account for the effect of $h$ in the objective function by multiplying the Euclidean distance heuristic by $\alpha / (1 - \phi_i + \epsilon)$. 

Using the A* path as a reference, we then solve the convex optimization problem formulated by \cite{ZhuSchmerlingEtAl2015} to produce a dynamically-feasible trajectory. This smoothing operation begins by placing a sequence of ``bubbles" about the points constituting the A* reference path defining a collision-free ``tube." Within this tube, the optimization problem is formulated as to minimize the curvature of the path subject to dynamic constraints. As an implementation consideration, we compute bubble radii for the A* path in the known region using the distance from the corresponding reference point to the nearest obstacle. Since we do not know the locations of the obstacles in the unknown regions, we select a constant, relatively small bubble radius\footnote{We have found values similar to the vehicle's turn radius to perform well as demonstrated in our experiments.} to ensure that the smoothed trajectory does not excessively stray from the reference path. 

Finally, the speed profile of this trajectory is optimized for minimum time traversal by solving the convex optimization problem presented by~\cite{LippBoyd2014}. In order to ensure safety, we enforce the additional constraint that the robot velocity must be zero at the end of $\sigma_\text{k}$. We reason that, provided the robot begins in a safe state and succeeds in planning a collision-free trajectory in the manner described, it will always be able to safely return to zero velocity within the known region of the environment. If any of the optimizations returns an infeasible solution at a particular iteration $i$, it would suffice to execute the remainder of the trajectory $\sigma_{\text{k}, i-1}$ planned at the previous iteration to safely bring the robot to rest.

\section{Experiments}\label{experiments}

We have trained a CNP model using data obtained from a set of 75 randomly-generated, single-path maze environments consisting of frequent corners and U-turns with 2.5 m hallways spanning a 25 m $\times$ 25 m area. To generate training data, we sample 504 unoccupied points from each map and simulate a 5 m range lidar scan. The spatial coordinates of the visible occupancy grid cells, expressed relative to the robot, and their occupancies then make up the context set $\mathcal{C}$. The target set $\mathcal{T}$ consists of the coordinates corresponding to points in $\mathcal{C}$ in addition to the coordinates of unobserved points within a 7.5 m radius about each frontier. Although our goal is to teach the model to predict the map beyond the frontiers, including observed points in $\mathcal{T}$ was found to benefit prediction accuracy. The model was trained for 1,000,000 iterations using a batch size of 4.

We evaluate our methodology\footnote{All of our source code, trained models, and data are publicly available online at \url{https://github.com/StanfordASL/MapPredictiveMotionPlanning}.} on a set of 30 randomly-generated mazes not seen during training. We compare against a na\"{i}ve frontier pursuit baseline and \cite{JansonHuEtAl2018}'s method. The na\"{i}ve baseline simply plans trajectories within the known map to the frontier centroid nearest the goal. Our implementation of \cite{JansonHuEtAl2018} faithfully recreates their heuristics, with two exceptions: (1) We use the method described in Section~\ref{MotionPlanningIIIB} for planning trajectories rather than FMT* (to reduce runtime complexity), and (2) we plan over an occupancy grid as opposed to a polygonal representation of the environment (to better match real-world hardware). 
Our CNP model was written in TensorFlow \cite{Abadi2015} with training and experimentation performed on a desktop computer running Ubuntu 18.04 equipped with an AMD Ryzen 1800X CPU and two NVIDIA GTX 1080 Ti GPUs. Path smoothing and trajectory optimization were performed with MOSEK \cite{MosekAPS2010} interfaced through Convex.jl \cite{UdellMohanEtAl2014} in Julia \cite{BezansonEdelmanEtAl2017}.
In this section, any stated $P$ values are obtained from two-tailed $t$-tests on the means of the two quantities differing.



\subsection{Planning with Map Predictions}

A key component of our methodology is the ability to predict what might lie in the unknown region, conditioned on current observations, and incorporate that prediction in motion planning. Fig.~\ref{pred_vis} shows a robot navigating through one of our mazes, visualizing our predictions of the unknown region. In it, we can see how our soft penalty encourages A* to plan paths through predicted free space for some distance into the unknown region, leading to smoothed paths that better prepare the robot for turns.

We can also see emergent behavior as a result of planning ahead with map predictions in Fig.~\ref{pred_vis}. Specifically, Fig.~\ref{pred_vis} (left) shows robot turning away from the second corner, prior to reaching it, enabling it to better hit the apex and maintain speed through the turn.

\subsection{Maze Completion Time Analysis}

  \begin{figure}[t]
    \centering
    \includegraphics[width=0.9\linewidth]{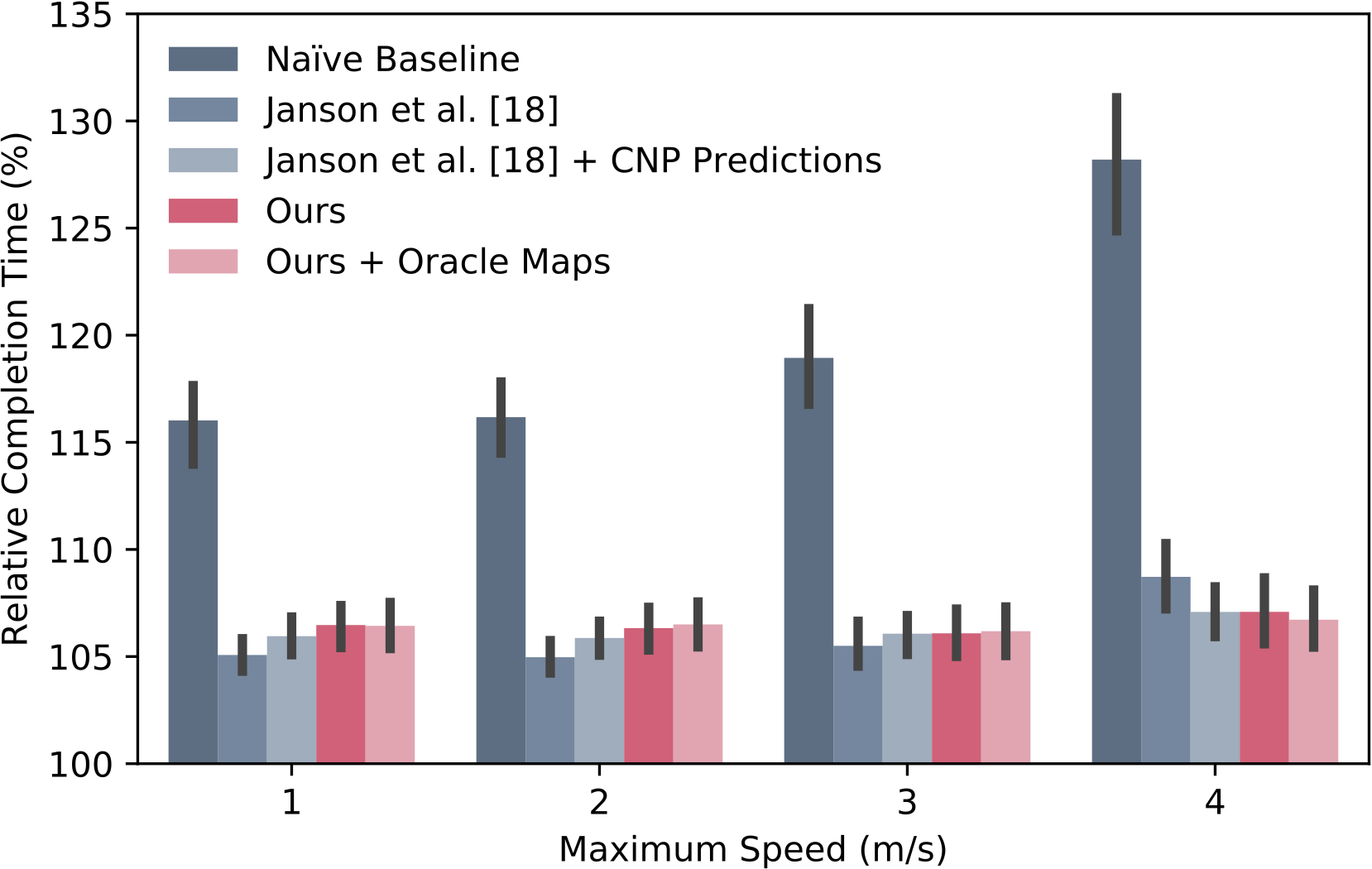}
    
    \vspace{-0.25cm}
    
    \caption{Mean completion times for each method relative to the optimal completion time (lower is better). Error bars show bootstrapped 95\% confidence intervals.}
    \label{bar_plot}
    \vspace{-0.5cm}
  \end{figure}

We quantitatively compare our method to the na\"{i}ve baseline and \cite{JansonHuEtAl2018} in terms of completion time relative to the optimal trajectory. In Fig.~\ref{bar_plot}, we can see that our method significantly outperforms the na\"{i}ve baseline ($P<.001$) for all maximum speed values. This is especially true as maximum speed increases, showing the efficacy of map predictions for maintaining path smoothness and speed through turns. Additionally, our method generally matches the performance of \cite{JansonHuEtAl2018}. For maximum speeds of 3 and 4 m/s, our method is directly comparable to \cite{JansonHuEtAl2018} ($P = \{0.17, 0.46\}$). For lower maximum speeds, our method is only 2\% worse in relative completion time (105\% vs.~107\%, $P = \{.042, .046\}$).

To determine if there is any performance that a better map prediction model could gain for our algorithm, we compare our method with an ideally-augmented version of itself (using perfect map predictions), which is referred to as ``Oracle Maps" in Fig.~\ref{bar_plot}. Our method performs equally as well as it would with perfect map predictions for all maximum speed values ($0.74 \leq P \leq 0.96$). This confirms that there is no performance to be gained by switching our CNP model for another type of map predictor.

Since our map predictions can be paired with any motion planner, we augmented \cite{JansonHuEtAl2018} with CNP predictions to determine how probabilistic map knowledge aids performance. As can be seen in Fig.~\ref{bar_plot}, the methods perform similarly ($0.12 \leq P \leq 0.43$). However, as maximum speed increases, \cite{JansonHuEtAl2018} with map predictions demonstrates more consistent performance compared to \cite{JansonHuEtAl2018}, even achieving a better average value at the highest maximum speed. This consistency in performance is also shared by our method as it maintains only a 7-8\% deficit to the optimal trajectory across maximum speed values, and is especially visible when compared to the na\"{i}ve baseline's worsening trend in Fig.~\ref{bar_plot}.





\subsection{Computation Time}

  \begin{figure}[t]
    \centering
    \includegraphics[width=0.9\linewidth]{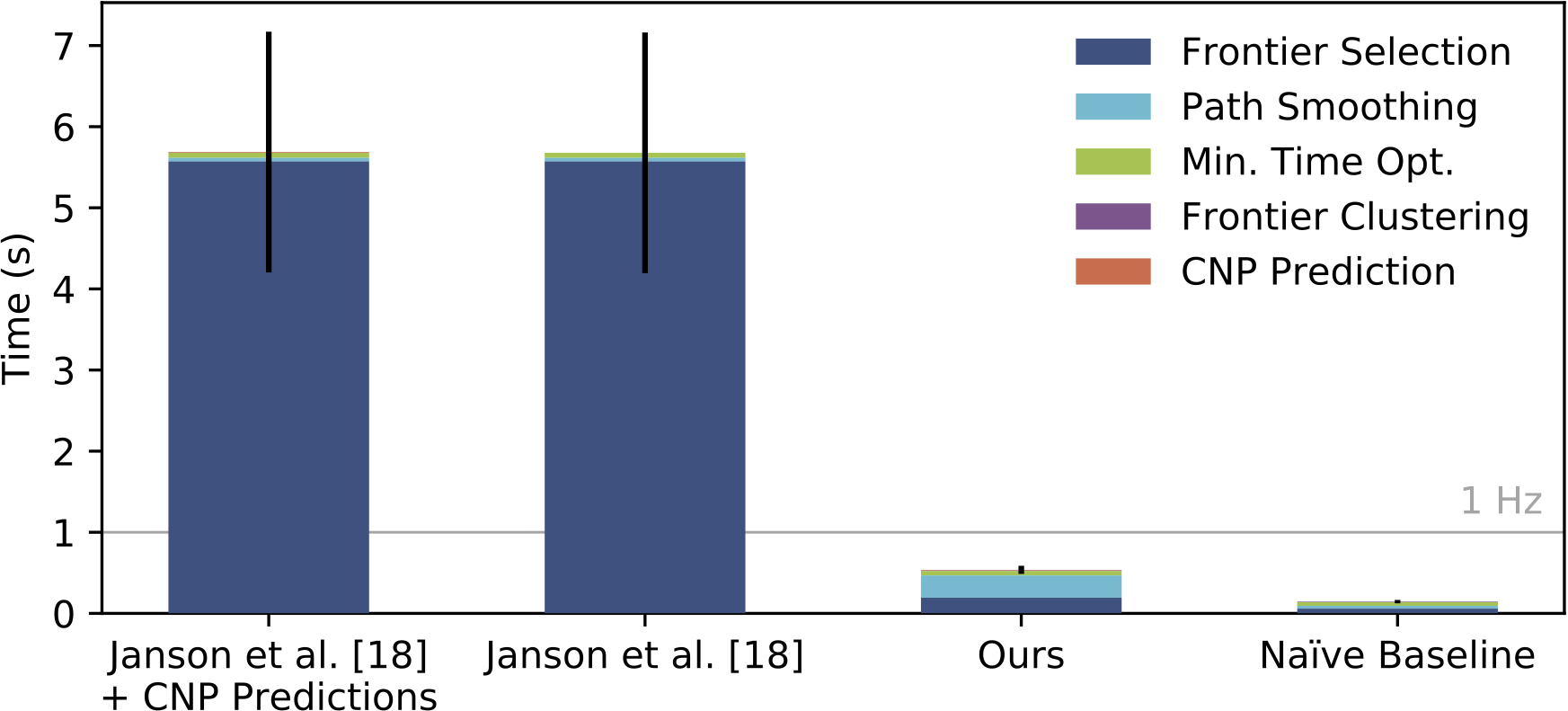}
    
    \vspace{-0.25cm}
    
    \caption{Mean runtime of one planning step for each method, with subcomponent contribution shown as proportions of each bar. Our method consistently runs at $\sim 2$ Hz. Error bars show 95\% confidence intervals.}
    \label{comp_time}
    \vspace{-0.4cm}
  \end{figure}


While our method and \cite{JansonHuEtAl2018} share similar performance in maze completion time, our method executes significantly faster. Fig.~\ref{comp_time} shows a comparison of the runtime of both methods, as well as \cite{JansonHuEtAl2018} with CNP predictions and the na\"{i}ve baseline. Our method, with an unoptimized Julia implementation and general-purpose solver, consistently achieves 0.5 s (2 Hz) execution time per planning iteration ($\sim 11 \times$ faster than \cite{JansonHuEtAl2018}). This is fast and predictable enough for real-time replanning on, e.g., a robot moving at 4 m/s with sensor and prediction ranges of 7.5 m and 5 m, respectively. The longest-running parts of our method are path smoothing and frontier selection, i.e., executing A* from the robot to the frontier centroids and from the frontier centroids to the goal, and smoothing the resulting path. Notably, obtaining map predictions by forward propagation of the CNP model yields negligable overhead, executing in 10 ms on average.

Beyond very long execution times, \cite{JansonHuEtAl2018} also suffers from a high runtime variance. This is undesirable in practice as it is difficult to estimate how far ahead to plan in order to ensure a real-time replanning frequency.

\subsection{Path and Speed Profiles}

  \begin{figure}[t]
    \centering
    \includegraphics[width=0.7\linewidth]{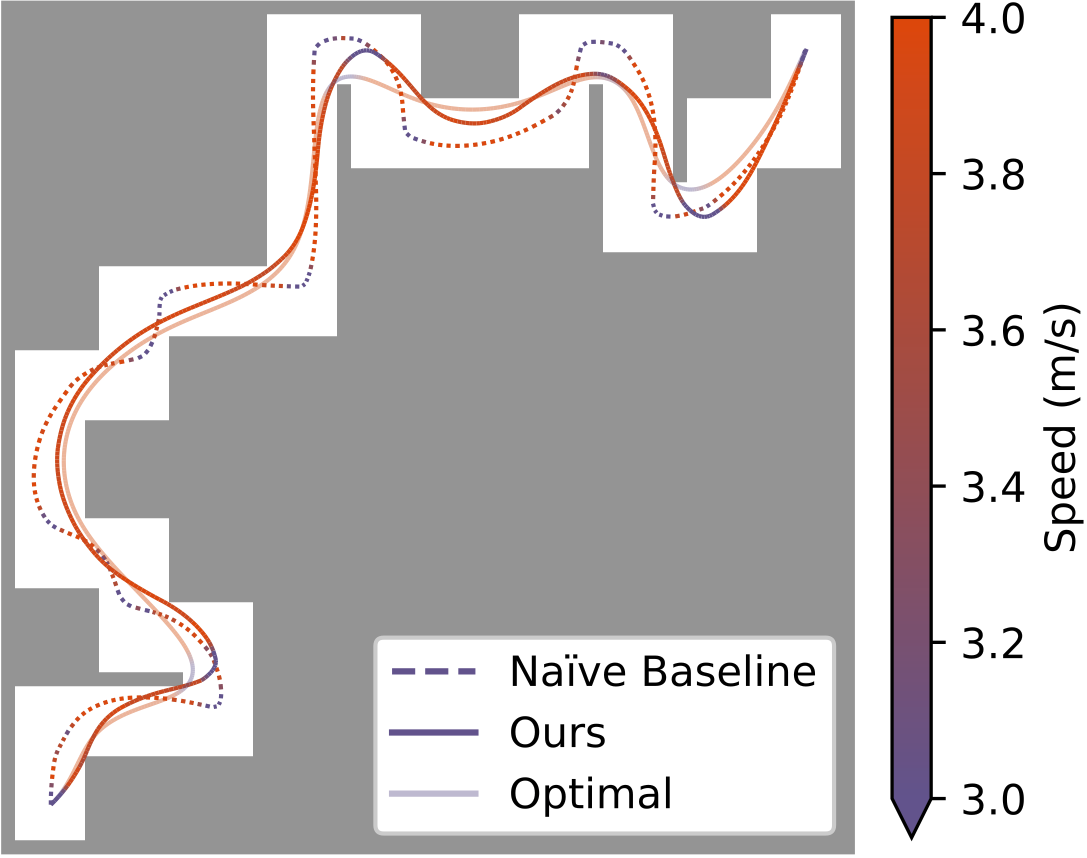}
    
    \vspace{-0.25cm}
    
    \caption{Our method, the na\"{i}ve baseline, and the optimal trajectory visualized with speed indicated by color. The baseline sees harsher braking and takes sharper turns than our method and the optimal trajectory.}
    \label{map_vis}
    \vspace{-0.2cm}
  \end{figure}
  
Fig.~\ref{map_vis} shows the trajectories and speed profiles obtained by the na\"{i}ve baseline, our method, and the optimal trajectory. As the baseline cannot foresee corners before they are visible, it acts late and needs to slow down heavily to make most turns. Conversely, our method closely follows the optimal trajectory and its speed profile through the map. This is especially noticeable in the first set of zig-zags where the baseline must slow down and turn for each corner. Our method predicts that the zig-zags will continue and plans a smooth, straight trajectory through them, nearly matching the optimal trajectory.

\subsection{Performance under Different Conditions}
  
  \begin{figure}[t]
    \centering
    \includegraphics[width=0.9\linewidth]{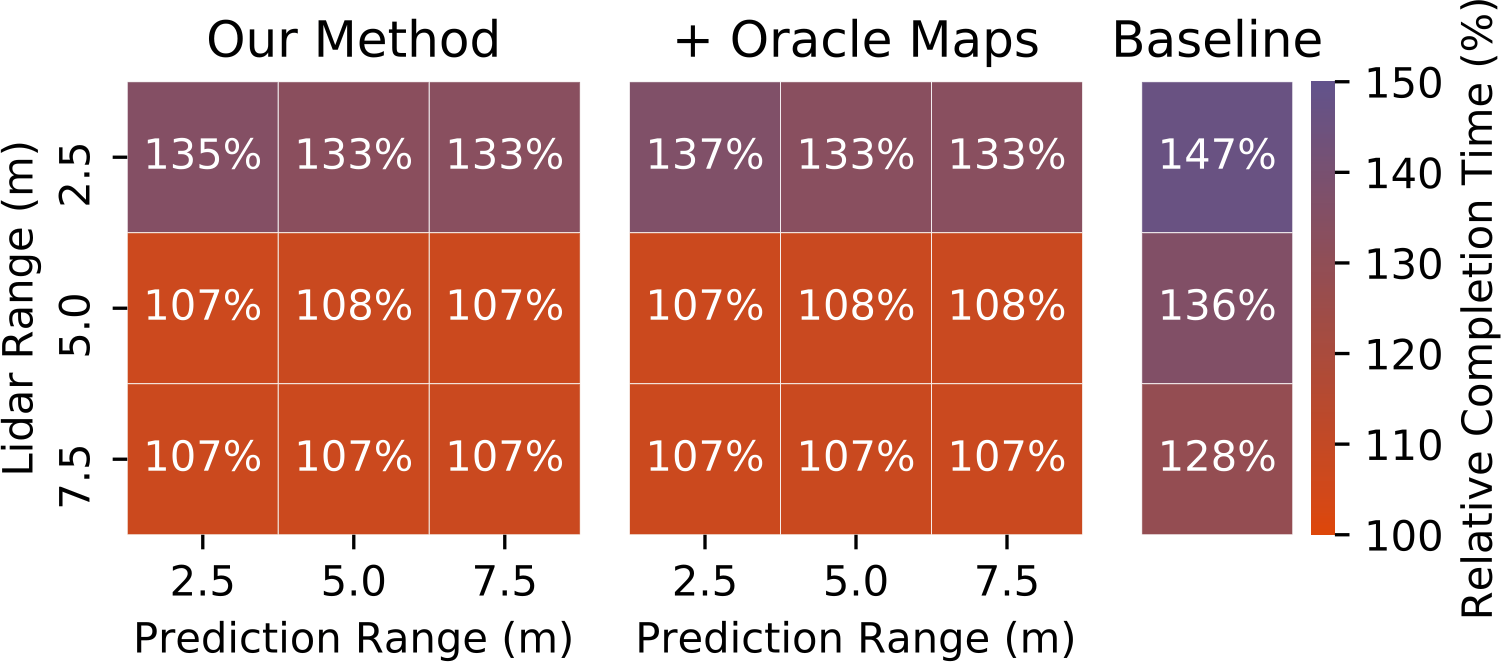}
    
    \vspace{-0.25cm}
    
    \caption{The relative completion times of our model, our model with perfect predictions, and the na\"{i}ve baseline for different sensor and prediction ranges with a maximum speed of 4 m/s.
    }
    \label{heatmap}
    \vspace{-0.5cm}
  \end{figure}
  
Since there are a variety of sensors used in robotics, there are many sensor ranges for which this method could be used. Our CNP was trained to predict 5 m into the unknown region from context points observed with a 7.5 m range laser scanner. To verify that the CNP can generalize to a variety of sensor and prediction ranges, we evaluate our method on different lidar and prediction ranges. Fig.~\ref{heatmap} shows the results of this evaluation.

As before, our method generally outperforms the na\"{i}ve baseline and matches the augmentation with ideal map predictions. Notably, all methods perform poorly when the sensor's range is 2.5 m.
Such a small observation range means the robot does not have enough time to accelerate to maximum speed before needing to slow down to maintain safety, degrading completion time. Importantly, we maintain algorithm performance for a prediction range of 7.5 m, which is larger than our CNP's trained prediction range of 5.0 m. 

\section{Conclusion}
In this work, we present a unified method that combines map prediction and motion planning to enable safe, time-efficient autonomous navigation of unknown environments with real-time performance. We demonstrate significant performance improvements in terms of time-to-goal compared to a na\"{i}ve frontier pursuit baseline. We also demonstrate significantly lower computation times than \cite{JansonHuEtAl2018}, which uses much better informed planning heuristics. Our method strikes a balance between performance and computational run-time constraints, while guaranteeing robot safety. Future work will evaluate the benefit of such predictions for motion planning in more realistic environments, e.g. real office environments. Further, we will deploy this algorithm on an RC car platform with dynamics matching those simulated in this work.

\addtolength{\textheight}{-11cm}




\clearpage
\newpage

\bibliographystyle{IEEEtran} 
\bibliography{ASL_papers,main}

\newcommand{\noopsort}[1]{} \newcommand{\printfirst}[2]{#1}
  \newcommand{\singleletter}[1]{#1} \newcommand{\switchargs}[2]{#2#1}
\begin{thebibliography}{10}
\providecommand{\url}[1]{#1}
\csname url@rmstyle\endcsname
\providecommand{\newblock}{\relax}
\providecommand{\bibinfo}[2]{#2}
\providecommand\BIBentrySTDinterwordspacing{\spaceskip=0pt\relax}
\providecommand\BIBentryALTinterwordstretchfactor{4}
\providecommand\BIBentryALTinterwordspacing{\spaceskip=\fontdimen2\font plus
\BIBentryALTinterwordstretchfactor\fontdimen3\font minus
  \fontdimen4\font\relax}
\providecommand\BIBforeignlanguage[2]{{%
\expandafter\ifx\csname l@#1\endcsname\relax
\typeout{** WARNING: IEEEtran.bst: No hyphenation pattern has been}%
\typeout{** loaded for the language `#1'. Using the pattern for}%
\typeout{** the default language instead.}%
\else
\language=\csname l@#1\endcsname
\fi
#2}}

\bibitem{BekrisKavraki2007}
K.~E. Bekris and L.~E. Kavraki, ``Greedy but safe replanning under kinodynamic
  constraints,'' in \emph{{Proc.\ IEEE Conf.\ on Robotics and Automation}},
  2007.

\bibitem{PivtoraikoMellingerEtAl2013}
M.~Pivtoraiko, D.~Mellinger, and V.~Kumar, ``Incremental micro-uav motion
  planning for exploring unknown environments,'' in \emph{{Proc.\ IEEE Conf.\
  on Robotics and Automation}}, 2013.

\bibitem{LiuWattersonEtAl2016}
S.~Liu, M.~Watterson, S.~Tang, and V.~Kumar, ``High speed navigation for
  quadrotors with limited onboard sensing,'' in \emph{{Proc.\ IEEE Conf.\ on
  Robotics and Automation}}, 2016.

\bibitem{Yamauchi1997}
B.~Yamauchi, ``A frontier-based approach for autonomous exploration,'' in
  \emph{{Proc.\ IEEE Int.\ Symp.\ on Computational Intelligence in Robotics and
  Automation}}, 1997.

\bibitem{BurgardMoorsEtAl2005}
W.~Burgard, M.~Moors, C.~Stachniss, and F.~E. Schneider, ``Coordinated
  multi-robot exploration,'' \emph{{IEEE Transactions on Robotics}}, vol.~21,
  no.~3, pp. 376--386, 2005.

\bibitem{SteinBradleyEtAl2018}
G.~J. Stein, C.~Bradley, and N.~Roy, ``Learning over subgoals for efficient
  navigation of structured, unknown environments,'' in \emph{{Conf.\ on Robot
  Learning}}, 2018.

\bibitem{CaleyLawranceEtAl2019}
J.~A. Caley, N.~R.~J. Lawrance, and G.~A. Hollinger, ``Deep learning of
  structured environments for robot search,'' \emph{{Autonomous Robots}},
  vol.~43, no.~7, pp. 1695--1714, 2019.

\bibitem{AydemirJensfeltEtAl2012}
A.~Aydemir, P.~Jensfelt, and J.~Folkesson, ``What can we learn from 38,000
  rooms? reasoning about unexplored space in indoor environments,'' in
  \emph{{IEEE/RSJ Int.\ Conf.\ on Intelligent Robots \& Systems}}, 2012.

\bibitem{ChangLeeEtAl2007b}
H.~J. Chang, C.~S.~G. Lee, Y.~Lu, and Y.~C. Hu, ``P-slam: Simultaneous
  localization and mapping with environmental-structure prediction,''
  \emph{{IEEE Transactions on Robotics}}, vol.~23, no.~2, pp. 281--293, 2007.

\bibitem{StroemNenciEtAl2015}
D.~P. Str{\"o}m, F.~Nenci, and C.~Stachniss, ``Predictive exploration
  considering previously mapped environments,'' in \emph{{Proc.\ IEEE Conf.\ on
  Robotics and Automation}}, 2015.

\bibitem{LiuRedaEtAl2018}
G.~Liu, F.~A. Reda, K.~J. Shih, T.~C. Wang, A.~Tao, and B.~Catanzaro, ``Image
  inpainting for irregular holes using partial convolutions,'' in
  \emph{{European Conf.\ on Computer Vision}}, 2018.

\bibitem{YuLinEtAl2018}
J.~Yu, Z.~Lin, J.~Yang, X.~Shen, X.~Lu, and T.~Huan, ``Generative image
  inpainting with contextual attention,'' in \emph{{IEEE Conf.\ on Computer
  Vision and Pattern Recognition}}, 2018.

\bibitem{ShresthaTianEtAl2019}
R.~Shrestha, F.~P. Tian, W.~Feng, P.~Tan, and R.~Vaughan, ``Learned map
  prediction for enhanced mobile robot exploration,'' in \emph{{Proc.\ IEEE
  Conf.\ on Robotics and Automation}}, 2019.

\bibitem{KatyalPopekEtAl2019}
K.~Katyal, K.~Popek, C.~Paxton, P.~Burlina, and G.~D. Hager,
  ``Uncertainty-aware occupancy map prediction using generative networks for
  robot navigation,'' in \emph{{Proc.\ IEEE Conf.\ on Robotics and
  Automation}}, 2019.

\bibitem{GuptaDavidsonEtAl2017}
S.~Gupta, J.~Davidson, S.~Levine, R.~Sukthankar, and J.~Malik, ``Cognitive
  mapping and planning for visual navigation,'' in \emph{{IEEE Conf.\ on
  Computer Vision and Pattern Recognition}}, 2017.

\bibitem{RichterVega2018}
C.~Richter, W.~Vega-Brown, and N.~Roy, ``Bayesian learning for safe high-speed
  navigation in unknown environments,'' in \emph{{Int.\ Journal of Robotics
  Research}}, 2018.

\bibitem{RichterRoy2017}
C.~Richter and N.~Roy, ``Safe visual navigation via deep learning and novelty
  detection,'' in \emph{{Robotics: Science and Systems}}, 2017.

\bibitem{JansonHuEtAl2018}
L.~Janson, T.~Hu, and M.~Pavone, ``Safe motion planning in unknown
  environments: Optimality benchmarks and tractable policies,'' in
  \emph{{Robotics: Science and Systems}}, 2018.

\bibitem{FraichardAsama2004}
T.~Fraichard and H.~Asama, ``Inevitable collision states -- a step towards
  safer robots?'' \emph{{Advanced Robotics}}, vol.~18, no.~10, pp. 1001--1024,
  2004.

\bibitem{GarneloRosenbaumEtAl2018}
M.~Garnelo, C.~M. Rosenbaum, D., T.~Ramalho, S.~M. Saxton, D., Y.~W. Teh,
  D.~Rezende, and S.~M.~A. Eslami, ``Conditional neural processes,'' in
  \emph{{Int.\ Conf.\ on Machine Learning}}, 2018.

\bibitem{JansonSchmerlingEtAl2015}
L.~Janson, E.~Schmerling, A.~Clark, and M.~Pavone, ``{Fast} {Marching} {Tree:}
  a fast marching sampling-based method for optimal motion planning in many
  dimensions,'' \emph{{Int.\ Journal of Robotics Research}}, vol.~34, no.~7,
  pp. 883--921, 2015.

\bibitem{ZhuSchmerlingEtAl2015}
Z.~Zhu, E.~Schmerling, and M.~Pavone, ``A convex optimization approach to
  smooth trajectories for motion planning with car-like robots,'' in
  \emph{{Proc.\ IEEE Conf.\ on Decision and Control}}, 2015.

\bibitem{LippBoyd2014}
T.~Lipp and S.~Boyd, ``Minimum-time speed optimisation over a fixed path,''
  \emph{{Int.\ Journal of Control}}, vol.~87, no.~6, pp. 1297--1311, 2014.

\bibitem{Abadi2015}
\BIBentryALTinterwordspacing
M.~Abadi \emph{et~al.} (2015) {TensorFlow}: Large-scale machine learning on
  heterogeneous systems. Software available from tensorflow.org. [Online].
  Available: \url{https://www.tensorflow.org/}
\BIBentrySTDinterwordspacing

\bibitem{MosekAPS2010}
{Mosek APS}. {The MOSEK optimization software}. {Available at
  }\url{http://www.mosek.com}.

\bibitem{UdellMohanEtAl2014}
M.~Udell, K.~Mohan, D.~Zeng, J.~Hong, S.~Diamond, and S.~Boyd, ``Convex
  optimization in {Julia},'' in \emph{{High Performance Technical Computing in
  Dynamic Languages}}, 2014.

\bibitem{BezansonEdelmanEtAl2017}
J.~Bezanson, A.~Edelman, S.~Karpinski, and V.~B. Shah, ``Julia: A fresh
  approach to numerical computing,'' \emph{{SIAM Review}}, vol.~59, no.~1, pp.
  65--98, 2017.

\end{thebibliography}

\end{document}